\documentclass[twoside]{article}

\usepackage{hyperref}
\usepackage{xcolor}
\hypersetup{
    colorlinks,
    linkcolor={blue!80!black},
    citecolor={blue!80!black},
    urlcolor={blue!80!black}
}
\usepackage{wrapfig}
\newcommand{\comment}[1]{}
\usepackage{multirow}

\usepackage{url}            % simple URL typesetting
\usepackage{booktabs}       % professional-quality tables
\usepackage{amsfonts}       % blackboard math symbols
\usepackage{nicefrac}       % compact symbols for 1/2, etc.
\usepackage{microtype}      % microtypography
\usepackage{amssymb,amsmath, amsthm, bm, cleveref}
\usepackage{algorithm}
\usepackage{algorithmic}
\usepackage[utf8]{inputenc}
\usepackage{mathtools}
\usepackage{subcaption}
\usepackage{tikz}
\usetikzlibrary{calc}
\usepackage{graphicx}
\usepackage{caption}
\usepackage{bbm}
\usepackage{bm}

\newtheorem{theorem}{Theorem}[section]

\newtheorem{assumption}{Assumption}[section]

\DeclareMathOperator*{\argmin}{arg\,min} % thin space, limits on side in displays
\DeclareMathOperator*{\argmax}{arg\,max} % thin space, limits on side in displays

\usepackage[accepted]{aistats2024}
\usepackage[round]{natbib}

\begin{document}
\newcommand{\ar}{\lambda}
\newcommand{\sr}{\mu}
\newcommand{\srv}{\bm{\mu}}
\newcommand{\sri}{\mu_i}
\newcommand{\buff}{l_M}
\newcommand{\nums}{k}
\newcommand{\busy}{B}
\newcommand{\ql}{L}
\newcommand{\q}{l}
\newcommand{\df}{\gamma}
\newcommand{\ts}{\theta}
\newcommand{\tv}{\bm{\theta}}
\newcommand{\ti}{\theta_i}
\newcommand{\g}{\sigma} % gradient / slope of the soft threshold
\newcommand{\cv}{\bm{\omega}} % critic parameters
\newcommand{\thrv}{\bm{m}}
\newcommand{\thr}{m}
\newcommand{\st}{s}
\newcommand{\stv}{\bm{s}}
\newcommand{\sti}{s_i}
\newcommand{\prob}{\Pr}
\newcommand{\pol}{\pi}
\newcommand{\dd}[1]{\mathrm{d}#1}
\newcommand{\ert}{T_r} % expected response time
\newcommand{\erti}{T_{r, i}}
\newcommand{\ssd}{\nu}
\newcommand{\val}{V}
\newcommand{\vf}{\bm{\phi}}
\newcommand{\pgName}{ACHQ}

\runningtitle{Efficient Reinforcement Learning for Routing Jobs in Heterogeneous Queueing Systems}

% \runningauthor{Surname 1, Surname 2, Surname 3, ...., Surname n}

\twocolumn[

\aistatstitle{Efficient Reinforcement Learning for Routing Jobs \\ in Heterogeneous Queueing Systems}

\aistatsauthor{ Neharika Jali \And Guannan Qu \And  Weina Wang \And Gauri Joshi }

\aistatsaddress{ Carnegie Mellon University 
% \And Carnegie Mellon University \And Carnegie Mellon University \And Carnegie Mellon University
} ]

\begin{abstract}
    We consider the problem of efficiently routing jobs that arrive into a central queue to a system of heterogeneous servers. Unlike homogeneous systems, a threshold policy, that routes jobs to the slow server(s) when the queue length exceeds a certain threshold, is known to be optimal for the one-fast-one-slow two-server system. But an optimal policy for the multi-server system is unknown and non-trivial to find. While Reinforcement Learning (RL) has been recognized to have great potential for learning policies in such cases, our problem has an exponentially large state space size, rendering standard RL inefficient. In this work, we propose \pgName, an efficient policy gradient based algorithm with a low dimensional soft threshold policy parameterization that leverages the underlying queueing structure. We provide stationary-point convergence guarantees for the general case and despite the low-dimensional parameterization prove that \pgName~converges to an approximate global optimum for the special case of two servers. Simulations demonstrate an improvement in expected response time of up to $\sim 30\%$ over the greedy policy that routes to the fastest available server.
\end{abstract}

\section{INTRODUCTION} \label{sec:intro}
\begin{figure}
    \centering    
    \includegraphics[width=\linewidth]{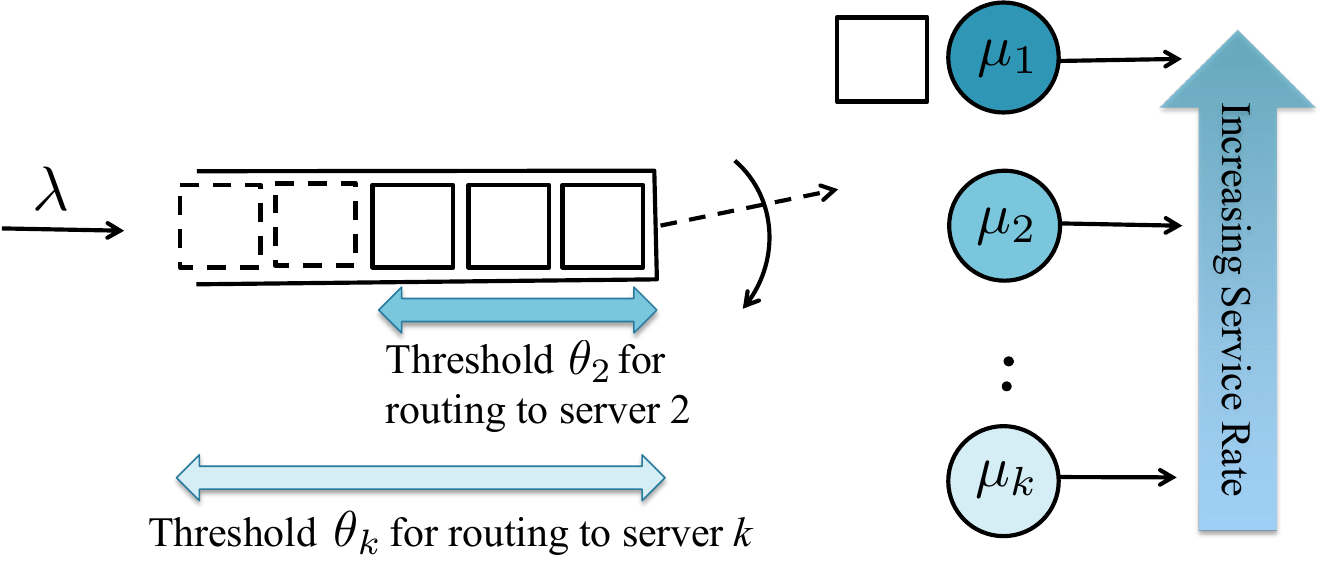}
    \caption{$k$ Server heterogeneous queueing system with service rates $\mu_i$, and job arrival rate $\lambda$. We also illustrate a threshold routing policy that routes jobs to a slower server $i$ when the queue length is $> \theta_i$.} 
   \vspace{-10pt}
    \label{fig:queueModel}
\end{figure}

With the recent exponential increase in large-scale cloud based services, we observe a paradigm shift in the nature of these systems and the way they serve jobs. A case in point is devices across generations of technology with varying capabilities present in the cluster. It is not uncommon to find a $\sim 4$ TFLOP K80, a $\sim 30$ TFLOP L4 and a $\sim 100$ TFLOP TPUv3 in the same facility \citep{googleGPU, googleTPU}. Another example is the differences in server speeds arising due to the fractional allocation of resources in serverless computing setups \citep{AWSLambda}. ML inference deployments with different servers hosting different sized models is yet another instance of varying service times \citep{shazeer2017outrageously}. Motivated by such heterogeneous systems, in this work, we consider the problem of efficient routing of jobs in queueing systems with servers of different speeds. We consider a stylized model with a single central queue and a system of heterogeneous servers as shown in \Cref{fig:queueModel}. Besides the cloud computing systems that motivated us, this model and the proposed routing policies are also applicable in other domains such as packet routing in communication networks \citep{srikant2013communication} or operations resource management in healthcare, manufacturing or ride-sharing \citep{walton2021learning, tsitsiklis2017flexible}. 

Traditional methods in queueing are often designed for homogeneous systems and do not account for the heterogeneity in service rates \citep{MorBook}. While these policies may still be efficient in particular load regimes or for certain types of heterogeneity, they are not optimal for the general case. For instance, a work-conserving routing policy that keeps all servers busy whenever there are jobs in the system is latency-optimal for homogeneous systems. However, it is sub-optimal for heterogeneous systems because it can be beneficial to keep slow servers idle and hold jobs in the central queue until the queue length exceeds a certain threshold or one of the fast server(s) becomes available. For the case of two servers - one fast and one slow - such a threshold policy has been shown to be optimal \citep{LinKumar, KooleVI, Walrand}. Despite many efforts, extending the proof of optimality of the threshold policy to a general multi-server case has been open for nearly four decades \citep{GerKooleOpinion}. Moreover, the threshold is a complicated and unknown function of the service rates and the job arrival rate. 

\paragraph{Main Contributions.} Motivated by difficulty in finding a closed-form expression for the policy, we present a reinforcement learning (RL) approach for finding the best routing policy in multi-server heterogeneous queueing systems. We make the following key contributions to address the inefficiency of off-the-shelf RL methods arising due to a large dimensional state space. (a) We leverage the underlying queueing structure to design a low-dimensional soft threshold policy parameterization and propose \pgName, an efficient policy gradient-based algorithm. (b) We provide stationary-point guarantees, and for the special case of two servers,  establish convergence to an approximate global optimum. (c) We demonstrate an improvement in expected response time of up to $\sim 30 \%$ over the greedy policy that routes to the fastest available server. To the best of our knowledge, ours is the first to propose an RL approach to this problem and one of few recent papers that leverage the queueing structure to design an efficient RL approach with provable guarantees.
 
%%%%%%%%%%%%%%%%%%%%%%

\section{RELATED WORK} \label{sec:relatedWork}
The problem of heterogeneous servers was first proposed by \cite{Larsen} in which the optimal policy was conjectured to be of threshold type. For two servers, a threshold policy was proved to be optimal using techniques of policy iteration \citep{LinKumar}, value iteration \citep{KooleVI} and sample path arguments \citep{Walrand}. While \cite{Rykov} claimed to have extended the proof to the general case of multiple heterogeneous servers, \cite{VericourtZhou} later showed that it was incomplete. The completeness and correctness of the proof in another attempt in \cite{Viniotis} has also been questioned by the authors of \cite{VericourtZhou}. Hence, despite several attempts, proving the optimality of a policy for the general multi-server case has been an open problem for nearly $40$ years \citep{GerKooleOpinion}.

Given the above challenges in deriving closed-form optimal policies, RL and approximate dynamic programming have been a natural choice for designing data-driven policies for queueing systems. Several works use these techniques effectively in applications such as scheduling in queueing networks \citep{VanRoy}, inventory control \citep{Mannor}, emergency response assignment \citep{VanBarneveld} and cooling in Google datacenters \citep{DeepMindCooling}. 

In recent years, there has been a renewed interest with a focus on large scale settings with possibly incomplete information and establishing guarantees of stability, convergence and correctness \citep{Ayesta, walton2021learning}. \cite{DaiGluzman} addresses the challenges of infinite state space, unbounded costs, and long-run average cost objective in queueing network control problems by proposing PPO and TRPO based deep RL algorithms with the policy optimization step enhanced by Lyapunov function arguments. \cite{Liu} proposes a truncation-based solution and establishes guarantees of optimality in the setting of unbounded state space by applying RL methods over a finite subset of the state space and a known stabilizing policy for the rest. \cite{QWI} focuses on developing intermediate solutions between model-free and model-based methods by exploiting the queueing structure to develop efficient learning algorithms. An alternate line of empirical work include \cite{StaffolaniRLQ} that implements a DoubleDQN method to learn task allocation in distributed queues and \cite{MaoDecima} that designs a REINFORCE based algorithm to learn workload specific efficient scheduling policies in distributed data processing jobs with dependency graphs. While further instances of RL in queues can be found in Section 5 of \cite{walton2021learning} and \cite{Ayesta}, to the best of our knowledge, our work is the first to study the multi-server heterogeneous problem from an RL perspective.

% Additionally, we include work that establish convergence guarantees for actor-critic policy gradient algorithms in Appendix C.

Reinforcement Learning approaches applied to large scale systems often suffer from state space explosion and standard algorithms are plagued by the curse of dimensionality. Actor Critic Policy Gradient offer a low dimensional solution by parameterizing the value function and policy. While actor-only methods are at a disadvantage due to high variance and inefficient use of samples between parameter updates and critic-only ones lack guarantees of optimality of resulting policy, actor critic methods provide the best of both worlds. We include works most relevant to our problem setting that establish guarantees of convergence below.

\cite{KondaAC} presents an asymptotic analysis of the average cost two timescale actor critic algorithm with a linear approximation of action-value function. A finite time convergence to a stationary point is established in \cite{Wu2022A2CPG} for the average cost two timescale advantage actor critic algorithm with a TD update of the critic and linear approximation of the state value function without assuming a compatible function approximation. Further, the authors also remark that their results may be extendable to the discounted setting using the same proof technique. \cite{KumarFinDisAC} analyze the finite time discounted cost actor critic algorithm with a Monte Carlo based update of the critic which is a linearly approximated action value function. A drawback of this paper is that it assumes that the action value function is indeed a linear function and can be approximated without error which might not be true for real world problems in general.

While there are a few papers that prove convergence to the global optimum for the special cases of linear MDPs and linear quadratic regulators, convergence for a general MDP is studied sparingly. \cite{pgALKM} proposes a discounted cost actor critic based Natural Policy Gradient algorithm with linear approximation of the action value function and compatible function approximation and establishes its global convergence. \cite{BhandariRussoGlobal} provide guarantees of optimality of a stationary point for a general policy gradient algorithm in the discounted setting when certain conditions of differentiability, closure and structure of the policy iteration objective are satisfied. To the best of our knowledge, demonstrating global convergence for the general MDP in the average cost setting is an open problem.

%%%%%%%%%%%%%%%%%%%%%%

\section{PROBLEM FORMULATION} \label{sec:problemSetup}
We first formally define the model and describe the characteristics of an optimal policy. We then model the problem as a Markov decision process. Finally, we demonstrate that standard RL methods suffer from the curse of dimensionality due to a large state space.
\subsection{Queueing Setup} \label{sec:queueModel}
\paragraph{Model.} We consider a queueing model with a single central queue and $\nums$ heterogeneous servers as shown in \Cref{fig:queueModel}. Jobs arrive into the queue according to a Poisson process with a known parameter $\ar > 0$. The queue is limited to hold at most $\buff$ jobs. A router then decides when and which server to route a job to according to a policy $\pi$. The time a job spends in service at server $i$ is modeled as an exponential random variable with a rate $\sri$ which is also known to the router. Without loss of generality, we assume that the servers are ordered as $\sr_1 \geq \sr_2 \geq \dots \geq \sr_{\nums}$. The system is stable when $\ar < \sum_{i=1}^{k} \sri$. We consider non-preemptive execution, where a job once routed to a server cannot be recalled or interrupted until it runs to completion.

\paragraph{Performance Metric.} The response time $\erti$ of the $i$-th job is defined as the time it takes from its arrival into the queue to its exit from the system after service which is a sum of its waiting and service times. The expected response time of the system following policy $\pol$ can be written as
\begin{equation} \label{eq:objective}
    \ert = \lim_{n \rightarrow \infty} \frac{1}{n} \mathbb{E}_{\pol} \left[\sum\limits_{i=0}^n \erti \right]
\end{equation}
where $\mathbb{E}_{\pol}$ denotes the expectation with respect to the stochastic process when policy $\pol$ is used. The objective, thus, is to find a policy that minimizes the expected response time $\ert$.

\paragraph{Slow Now versus Fast Later.} If the goal were to maximize throughput instead of average response time, the problem becomes trivial and the optimal policy would be to just keep all the servers busy. For our objective \eqref{eq:objective}, a greedy policy that always routes to the fastest among the available servers is not optimal because faster server(s) can become available soon after the router assigns all jobs in the queue to slower server(s). Thus, the goal of a policy optimizing the expected response time is to balance the trade-off between sending a job right now to a slower server, or waiting for a faster server to become available. 

\paragraph{Threshold Policy.} As depicted in \Cref{fig:queueModel}, a threshold policy is a rule that chooses to route a job to the fastest available server (say server $i$) when the number of jobs waiting in the queue exceeds a threshold $\ts_i$ and wait for the next timestep otherwise. Note that the threshold $\theta_1 = 0$, since it is optimal to route jobs to the fastest server whenever it is available. Such a policy captures the intuition that if there are a lot of jobs in the queue and there is pressure to clear the backlog, the router should send a job to a slower server right now. But if there are only a few jobs in the queue, then it should wait for a faster server to become available. For a queueing system with only two servers, one fast and one slow, a threshold policy has been proved to be optimal \citep{LinKumar, KooleVI, Walrand}. In this work, we consider the more general $\nums$ server case, which has been an open problem for nearly four decades \citep{GerKooleOpinion}. 

\subsection{Queue as a Markov Decision Process} \label{sec:queueAsMDP}
The Markov chain view of the queue renders utilizing tools from reinforcement learning a natural choice. We model the queue as an average cost discrete-time Markov decision process (MDP) $\mathcal{M} = (\mathcal{S}, \mathcal{A}, \mathcal{P}, c)$ where $\mathcal{S}, \mathcal{A}$ are the state and action spaces respectively, $\mathcal{P}$ the transition probability and $c$ the cost. The continuous time queue described in the previous subsection can be converted to an equivalent discrete-time representation by sampling the systems at instants of arrivals and departures \citep{LippmanUniformization}. Hence, we have a discrete-time system, where at each timestep, whose beginning is marked by an arrival or a departure, the router decides if and to which server to send a job waiting in the queue.

The state of the system at time $t$ is represented as $\stv^{(t)} = (\ql^{(t)}, \busy_1^{(t)}, \dots, \busy_{\nums}^{(t)})$ where $\ql^{(t)} \in [\buff]$ is the number of jobs in the queue and $\busy_i^{(t)} \in \{0, 1\}$ represents if server $i$ is available or busy respectively. An action $a \in \mathcal{A}$ corresponds to routing a job to an idle server $i$, $a = i$, or keeping the job in the queue and choosing to wait for the next time step, $a = 0$. Notice that the state space scales exponentially in the number of servers $\lvert \mathcal{S} \rvert = (\buff + 1) 2^\nums$ and algorithms that depend on the size of the state space suffer from the curse of dimensionality. 

State transitions occur whenever there is an arrival or a departure and the corresponding transition probabilities $\prob(\stv' | \stv, a)$ are a function of the arrival $\ar$ and service rates $\srv = (\sr_1, \dots, \sr_{\nums})$ and the action taken. Note that the finite Markov chain under consideration is irreducible and aperiodic, which implies ergodicity. For ergodic systems,  by Little's Law, the objective of minimizing the expected response time is equivalent to minimizing the time-averaged number of jobs in the system (including both those waiting in queue and those being currently served) \citep{MorBook}. We, hence, define the cost as the total number of jobs in the system $c(\stv) = \sum_i \sti$ $= \ql + \sum_j \busy_j$, where $\sti$ denotes the $i$-th element of $\stv$. The policy is represented as $\pol(\cdot|\stv)$. Denote the stationary distribution over the states induced by the policy as $\ssd_\pol$. 

The average cost is defined as 
\begin{equation}
    c_{\pol} := \lim\limits_{T \rightarrow \infty} \frac{\sum_{t=0}^T c(\stv^{(t)})}{T} = \mathbb{E}_{\stv \sim \ssd_{\pol}} [c(\stv)].
\end{equation}
Further, we also define the overall cost accumulated when starting from state $\stv^{(0)}$ and following policy $\pol$ by the state-value function 
\begin{equation*}
    \val^{\pol}(\stv) := \mathbb{E} \left[\sum\limits_{t=0}^\infty (c(\stv^{(t)}) - c_{\pol} ) \lvert \stv^{(0)} = \stv \right].
\end{equation*}
Another useful quantity is the action-value function
\begin{equation*}
    Q^{\pol}(\stv, a) := \mathbb{E} \left[\sum\limits_{t=0}^\infty (c(\stv^{(t)}) - c_{\pol} ) \lvert \stv^{(0)} = \stv, a^{(0)} = a\right].
\end{equation*}

The goal thus is to find the optimal policy $\pol^\star(\cdot |\stv)$ that minimizes the average cost, $\pol^\star = \argmin\limits_{\pol} c_{\pol}$.

\subsection{Relative Value Iteration: \hspace{100pt} The Curse of Dimensionality} \label{sec:algoRVI}
%
% Recall that while for the two server case, a policy of threshold type is known to optimal, there is no analytical expression for calculating the threshold. Further, 
Recall that for the multi-server case, the optimal policy is unknown. Relative Value Iteration (RVI) which is the Value Iteration algorithm adapted to average cost \citep{WhiteRVI} addresses this. We include a brief description below and pseudo-code in Appendix C.

Define the Bellman operator over the value function as $$T(\val)(\stv) = \min\limits_{a \in \mathcal{A}} c(\stv) + \sum\limits_{\stv' \in \mathcal{S}} \prob(\stv' \lvert \stv, a) \val(\stv').$$ Value Iteration computes $\val(\stv)$ by iteratively updating values for every state using the Bellman equation in every round. Additionally, in the average cost case, RVI subtracts the value of a reference state  $\stv^\star \in \mathcal{S}$, from the values of other states to mitigate the numerical instability caused by large value functions \citep{WhiteRVI} $$\val^{(t+1)}(\stv) \gets T(\val^{(t)})(\stv) - T(\val^{(t)})(\stv^\star).$$

While it is convenient to use RVI when there are a limited number of servers and a small buffer, it becomes computationally intractable in large scale real-world applications. As the state space scales exponentially in the number of servers, the complexity of each iteration of value function updates also scales exponentially. For example, there are $20,480$ states in a system with $\nums = 10$ servers and a buffer capacity of $\buff = 20$. To address this issue, we leverage the underlying structure in the queueing model and present a function approximation-based policy gradient algorithm in the next section.

%%%%%%%%%%%%%%%%%%%%%%

\section{ACTOR CRITIC FOR HETEROGENEOUS QUEUES} \label{sec:algoPG}
In this section, we present \pgName~-- Actor Critic for Heterogeneous Queues. \Cref{algo:PG} is a policy gradient method with function approximation that leverages threshold-like-policy properties of the queueing system. We first briefly outline the two time-scale Actor Critic algorithm \citep{SuttonBartoBook} and then describe the parameterization that we design to mitigate the curse of dimensionality.

\begin{algorithm}
    \caption{\pgName} \label{algo:PG}
    \begin{algorithmic}[1]
         \STATE \textbf{Initialize} actor parameters $\tv^{(0)}$; critic parameters $\cv^{(0)}$; average cost estimator $\eta^{(0)}$; step sizes $\alpha^{(t)}$ for actor, $\beta^{(t)}$ for critic, $\zeta^{(t)}$ for average cost estimator
         \STATE Draw $\stv^{(0)}$ from some initial distribution 
         \FOR{$t = 0, 1, 2, \dots$}
            \STATE Take action $a^{(t)} \sim \pol_{\tv^{(t)}}(\cdot \lvert \stv^{(t)})$
            \STATE Observe cost $c^{(t)} = c(\stv^{(t)})$
            \STATE Observe next state $\stv^{(t+1)} \sim \prob(\cdot \lvert \stv^{(t)}, a^{(t)})$
            \STATE $\delta^{(t)} = c^{(t)} - \eta^{(t)} + \vf(\stv^{(t+1)})^T \cv^{(t)} - \vf(\stv^{(t)})^T\cv^{(t)}$ \label{line:advantage}
            \STATE $\eta^{(t+1)} = \eta^{(t)} + \zeta^{(t)}(c^{(t)} - \eta^{(t)})$ \label{line:averageReward}
            \STATE $\cv^{(t+1)} = \Pi_{R_{\cv}}\left(\cv^{(t)} + \beta^{(t)}\delta^{(t)}\vf(\stv^{(t)})\right)$ \label{line:criticParameters}
            \STATE $\tv^{(t+1)} = \tv^{(t)} - \alpha^{(t)}\delta^{(t)}\nabla_{\tv}\log \pol_{\tv^{(t)}} (a^{(t)} \lvert \stv^{(t)})$ \label{line:actorParameters}
         \ENDFOR
    \end{algorithmic}
\end{algorithm}

 \subsection{Preliminaries}
Consider the policy $\pol_{\tv}(\cdot | s)$, also known as the \emph{actor}, to be parameterized by $\tv$. Denote the stationary distribution induced over the states by this policy by $\ssd_{\tv}$, value function by $\val^{\tv}$ and action-value function by $Q^{\tv}$. The performance of $\pol_{\tv}$ is measured by the expected cost under the stationary distribution $\ssd_{\tv}$ which is 
\begin{equation}
    J(\tv) := c_{\pol_{\tv}} = \mathbb{E}_{\stv \sim \ssd_{\tv}}[c(\stv)].
\end{equation}
%
% The algorithm finds the optimal in the class of policies parameterized by $\tv$ as $\tv^\star = \argmin_{\tv} J(\tv)$ using gradient descent. Using the Policy Gradient Theorem and a baseline for variance reduction, we have $$\nabla J(\tv) = \mathbb{E}_{\stv, a} [\Delta^{\tv}(\stv, a) \nabla \log \pol_{\tv}(\stv, a)]$$ where $\Delta^{\tv}(\stv, a) := Q^{\tv}(\stv, a) - \val^{\tv}(\stv)$ is the \emph{advantage}. We model the \emph{critic} to be approximated by a linear function to mitigate the exponential state space problem as 
% \begin{equation}
%     \Hat{V}(\stv; \cv) = \vf(\stv)^T \cv.
% \end{equation}
% Putting together the actor and critic and using a TD(0) update for the critic parameters, we have \Cref{algo:PG}. Note that $\Pi_{R_{\cv}}$ represents the projection to appropriately chosen radius $R_{\cv}$
%
\paragraph{Actor.} The algorithm finds the optimal in the class of policies parameterized by $\tv$ as $\tv^\star = \argmin_{\tv} J(\tv)$ using gradient descent which can be represented as
\begin{equation*}
    \tv^{(t+1)} = \tv^{(t)} - \alpha \nabla J(\tv^{(t)}).
\end{equation*}
By the Policy Gradient theorem, we can represent $\nabla J(\tv) = \mathbb{E}_{\stv \sim \ssd_{\tv}, a \sim \pol_{\tv}} [Q^{\tv}(\stv, a) \nabla \log \pol_{\tv}(\stv, a)]$ \citep{SuttonBartoBook}. In practice, a baseline (such as $\val$) is subtracted from the action-value function $Q$ while estimating gradients to reduce variance which is as follows
\begin{equation*}
    \nabla J(\tv) = \mathbb{E}_{\stv, a} [\Delta^{\tv}(\stv, a) \nabla \log \pol_{\tv}(\stv, a)]
\end{equation*}
where $\Delta^{\tv}(\stv, a) := Q^{\tv}(\stv, a) - \val^{\tv}(\stv)$ is the \emph{advantage}. 

\paragraph{Critic.} To mitigate the exponential state space problem, we model the critic to be approximated by a linear function
\begin{equation}
    \Hat{V}(\stv; \cv) = \vf(\stv)^T \cv.
\end{equation}
The gradient step for the critic can be represented as 
\begin{equation*}
    \cv^{(t+1)} = \cv^{(t)} + \beta(V(\stv^{(t)}) - \vf(\stv^{(t)})^T\cv)\vf(\stv^{(t)}).
\end{equation*}

\paragraph{Algorithm.} Putting together the actor and critic, as described above, we have \Cref{algo:PG}. Expression in line~\ref{line:advantage} is a result of the advantage estimated as 
\begin{align}
    \delta^{(t)} & = \Hat{Q}(\stv^{(t)}, a^{(t)}) - \Hat{V}(\stv^{(t)}) \nonumber \\
    & = c^{(t)} - \eta^{(t)} + \hat{V}(\stv^{(t+1)}) - \hat{V}(\stv^{(t)}) \nonumber
\end{align}
where $\eta^{(t)}$ denotes the average cost until timestep $t$. This estimate is then used in the gradient step for the actor parameters in line~\ref{line:actorParameters}. Using a TD(0) style update, for the critic parameters results in the expression in line~\ref{line:criticParameters} where $\Pi_{R_{\cv}}$ represents the projection to appropriately chosen radius $R_{\cv}$.
% $$\hat{V}(\stv^{(t)}) = c^{(t)} - \eta^{(t)} + \hat{V}(\stv^{(t+1)}),$$

\subsection{Low Dimensional Parameter Design}
Recall that the number of states in our problem scales exponentially in the number of servers. To alleviate this issue of a very large state space, we design a low dimensional actor and critic leveraging properties of the queueing system. A threshold policy, as defined in \Cref{sec:queueModel}, is known to be optimal for the two server case and is conjectured to be optimal for the multi-server system \citep{LinKumar}. Moreover, we observe the optimal policy to be of threshold type in simulations in \Cref{sec:expRVI} for the multi-server model.

\paragraph{Policy Parameters.} First, we note that if the fastest server is free and there is a job waiting in the queue, then the job is routed to the fastest server irrespective of the number of jobs waiting in the queue. This can be argued because the job can be served no faster than by the fastest server and hence there is no incentive to wait. We then parameterize the actor by $\tv \in \mathbb{R}^{\nums-1}$ as a soft threshold policy with one threshold per server for servers $2, \dots, \nums$. If $f = \argmax_i \sri (1-\busy_i)$ is the index of the fastest among the available servers in state $\stv$, $\ts_f$ the threshold corresponding to it, $\st_0$ the number of jobs in the queue and $\g$ a hyperparameter that controls the slope or sharpness of the decision boundary, then the probability of routing a job to server $f$ is
\begin{align} \label{eqn:softThresholdPolicy}
    \pol_{\tv}(a = f \lvert \stv) &= \frac{e^{\g (\st_0 - \ts_f)}}{1 + e^{\g (\st_0 - \ts_f)}}.
    % \pol_{\tv}(a = 0 \lvert \stv) & = \frac{1}{1 + e^{\g (\st_0 - \ts_f)}}
\end{align}
Further, the probability of waiting until the next timestep is $\pol_{\tv}(a = 0 \lvert \stv) = 1 - \pol_{\tv}(a = f \lvert \stv)$. This choice of parameterization is a soft, differentiable version of the threshold policy conjectured to be optimal in \cite{LinKumar}.

\paragraph{Value Function Approximation.} We choose the features for linear approximation of the value function
\begin{equation} \label{eqn:vf}
    \vf(\stv) = \frac{\stv}{\buff + \nums} .  
\end{equation}
This reflects the intuition that the more the number of jobs in the system, the longer it takes to serve them and hence higher is the cost. Note that the feature vector is normalized by $\buff + \nums$ to ensure that $\lvert \lvert \vf(\cdot) \rvert\rvert < 1$.

\section{CONVERGENCE GUARANTEES} \label{sec:theory}
We first show that \pgName~converges to a stationary point for the multi-server case by applying results from \cite{Wu2022A2CPG}. We then prove that the stationary point is an approximate global optimum for the special case of two servers in the discounted cost setting by applying results from \cite{BhandariRussoGlobal}.

\subsection{Convergence to Stationary Point} \label{sec:theoryStationaryPoint}
In this subsection, we show that the soft threshold policy defined in \Cref{eqn:softThresholdPolicy} converges to a first order stationary point by applying the results of finite time two timescale actor critic methods from \cite{Wu2022A2CPG}. We detail the assumptions considered, verify that they hold for our problem and then state the convergence rate and sample complexity for \Cref{algo:PG}.

\begin{assumption}[Bounded Feature Norm] \label{asm:boundedFeatureNorm}
    The norm of the feature vector is bounded i.e. $\lvert \lvert \vf(\cdot) \rvert\rvert < 1$.
\end{assumption}
% This is ensured by design in \Cref{eqn:vf}.

\begin{assumption}[Negative Definite $\bm{A}$] \label{asm:negDef}
    For all policy parameters $\tv$, $\bm{A} := \mathbb{E}_{\stv, a, \stv'}[\vf(\stv)(\vf(\stv') - \vf(\stv))^T]$ is negative definite.
\end{assumption}

\begin{assumption}[Uniform Ergodicty] \label{asm:uErg}
    Consider the Markov chain generated by the rule $a^{(t)} \sim \pol_{\tv}(\cdot \lvert \stv^{(t)})$, $\stv^{(t+1)} \sim \prob(\cdot \lvert \stv^{(t)}, a^{(t)})$ and the stationary distribution $\ssd_{\tv}$ induced by policy $\pol_{\tv}$. There exists an $m > 0$ and $\chi \in (0, 1)$ $\forall \tau \geq 0, \forall \stv \in \mathcal{S}$ such that
    \begin{equation*}
        d_{TV}(\prob(\stv^{(\tau)} \in \cdot \lvert \stv^{(0)} = \stv), \ssd_{\tv}(\cdot)) \leq m \chi^\tau.
    \end{equation*}
\end{assumption}

\begin{assumption}[Bounded and Lipschitz Continuous Gradients and Policy] \label{asm:cL}
    There exists constants $y_1, y_2, y_3 > 0$ such that for all given states $\stv$, actions $a$ and $\forall \tv \in \mathbb{R}^\nums$
    \begin{align*}
        & ||\nabla \log \pol_{\tv}(a | \stv)|| \leq y_1, \\
        & ||\nabla \log \pol_{\tv_1} (a | \stv) - \nabla \log \pol_{\tv_2} (a | \stv)|| \leq y_2 ||\tv_1 - \tv_2||, \\
        & |\pi_{\tv_1} (a | \stv) - \pi_{\tv_2} (a | \stv) | \leq y_3 ||\tv_1 - \tv_2||.
    \end{align*}
\end{assumption}

\begin{assumption} [Bounded Value Function Approximation Error] \label{asm:vApprox}
    The value function approximation error
    \begin{equation*}
        \epsilon_{app}(\tv) := \sqrt{\mathbb{E}_{\stv \sim \ssd_{\tv}}\left[(\vf(\stv)^T \cv^\star(\tv) - \val^{\tv}(\stv))^2\right]}
    \end{equation*}
    is uniformly bounded for all potential policies by some constant $\epsilon_{app} \geq 0$, $\forall \tv$ as $\epsilon_{app}(\tv) \leq \epsilon_{app}$.
\end{assumption}

\paragraph{Assumption Verification.} Assumption~\ref{asm:boundedFeatureNorm} is ensured by design in \Cref{eqn:vf}. Assumption~\ref{asm:negDef} is equivalent to $\Phi$, the matrix whose columns are $\vf(\stv)$, being full rank \citep{ZhangAvgRewardTD}. This condition is satisfied in our case due to the presence of $\bm{e}_i$ among columns of $\Phi$ where $e_i$ is a vector with all but $i$-th element as zeros and $i$-th element as $1/(\buff + \nums)$. Assumption~\ref{asm:uErg} is fulfilled due to the irreducibility and aperiodicity of our finite Markov chain \citep{Bhandari2018finite}. It is easy to verify that properties in Assumption~\ref{asm:cL} hold for our policy parameterization in \Cref{eqn:softThresholdPolicy}. Satisfaction of Assumption~\ref{asm:vApprox} is demonstrated by experiments in \Cref{sec:expRVI} that show the value function is approximately linear in the state vector. Note that proving the value function to be linear in the state vector is a hard problem since the optimal policy is unknown for the multi-server system. 

We now formally state the convergence rate and sample complexity. Note that this result follows from Theorem 4.5, 4.7 of \cite{Wu2022A2CPG} that establish the convergence of the actor and critic respectively.
\begin{theorem}[\cite{Wu2022A2CPG}, Corollary~4.9]
    Under assumptions 5.1-5.5, choosing the actor step size $\alpha^{(t)} = \mathcal{O}(1/(1+t)^{r_\alpha})$ and the critic step size $\beta^{(t)} = \mathcal{O}(1/(1+t)^{r_\beta})$, where $0 < r_\beta < r_\alpha < 1$, we have 
    \begin{align}
        \min\limits_{0 \leq i \leq t} \mathbb{E}||\nabla J(\tv^{(i)})||^2 & = \mathcal{O}(\epsilon_{app}) + \mathcal{O}\left(\frac{1}{t^{1-r_\alpha}}\right) \nonumber \\
        & + \mathcal{O}\left(\frac{\log t}{t^{r_\beta}}\right) + \mathcal{O}\left(\frac{1}{t^{2(r_{\alpha} - r_{\beta})}}\right).
    \end{align}
    \Cref{algo:PG} can find an $\epsilon$-approximate stationary point of $J(\cdot)$ within $\tau$ steps as
    \begin{equation}
        \min\limits_{0 \leq i \leq \tau} \mathbb{E}||\nabla J(\tv^{(i)})||^2 \leq \mathcal{O}(\epsilon_{app}) + \epsilon,
    \end{equation}
    where $r_{\alpha} = 3/5, r_{\beta} = 2/5$ and the total number of iterations $\tau = \Tilde{\mathcal{O}}(\epsilon^{-2.5})$. 
\end{theorem}

\subsection{Approximate Optimality of Stationary Point for Two Servers} \label{sec:theoryTwoServers}

While we proved that \pgName~converges to a stationary point in the previous section, we are yet to establish global optimality of such a stationary point. Moreover, it is unknown whether the optimal policy even lies in the class of threshold policies. Recall that finding an optimal policy for the general multi-server case is an open problem in queueing theory. Here, we consider the special case of two servers and a discounted cost MDP. We show that any stationary point is approximately optimal by using results from \cite{BhandariRussoGlobal}. We define some preliminaries, describe the required conditions, show that they hold for our queueing model and finally state the formal result.

Here we look at the special case of two servers $\nums = 2$ - one fast and one slow. Recall that the optimal policy is proven to be of threshold type for this case. Further, we consider the discounted cost MDP here. While we consider the average cost MDP in \Cref{algo:PG} and the guarantees of convergence to a stationary point in \Cref{sec:theoryStationaryPoint}, we note that a global optimality analysis for the average cost MDPs is still an open problem. We also observe empirically in \Cref{sec:expPG}, a similar performance (and often with fewer samples) with a discounted cost. We continue to use the soft threshold policy parameterization described in \Cref{eqn:softThresholdPolicy} and represent by $\Theta$ the class of such policies.

The state-value function $\val^{\pol}_{\df}(\stv)$ is now defined as the overall discounted cost accumulated when starting from state $\stv$ and following policy $\pol$
\begin{equation*}
    \val^{\pol}_{\df}(\stv) := \mathbb{E}\left[\sum\limits_{t=0}^{\infty} \df^t c(\stv^{(t)}) \lvert \stv^{(0)} = \stv \right]
\end{equation*}
where $\df$ is the discount factor. Represent the distribution from which the initial state is chosen by $\xi$. The performance measure can hence be written as 
\begin{equation*}
    J_{\df}(\pol) = \mathbb{E}_{\stv \sim \xi} \left[\val^{\pol}(\stv)\right]
\end{equation*}
and the optimal policy is $\pol^\star = \argmin_{\pol} J_{\df}(\pol)$. 

Define the weighted Policy Iteration (PI) objective or the "Bellman" cost function as 
\begin{align}
    \mathcal{B}\left(\Bar{\pol} \lvert \ssd, \val^{\pol}_\df \right) & := \mathbb{E}_{\stv \sim \ssd} \left[(T_{\Bar{\pol}} \val^{\pol}_\df)(\stv)\right] \nonumber \\ 
    & = \mathbb{E}\left[Q^{\pol}_\df(\stv, \Bar{\pol}(\stv))\right]
\end{align}
for a probability distribution $\ssd$ over state space $\mathcal{S}$ where $T_{\Bar{\pol}}$ is the Bellman operator with respect to policy $\Bar{\pol}$. Further, define the effective concentrability coefficient $\kappa_{\xi}$ for the class of value functions $\mathcal{V}_{\Theta} = \{\val^{\pol_{\ts}} : \ts \in \Theta \}$ to be the smallest scalar such that $\forall \val \in \mathcal{\val}_{\Theta}$
\begin{equation*}
    ||\val - \val^\star||_{1, \xi} \leq \frac{\kappa_{\xi}}{(1-\df)} ||\val - T\val||_{1, \xi}
\end{equation*}
where $||\val(\stv)||_{1, \xi} = \mathbb{E}_{\stv \sim \xi}[\val(\stv)]$.

\begin{assumption}[Differentiability] \label{asm:diff}
    For each policy parameter $\ts$ and induced stationary distribution $\ssd_{\ts}$, the functions $\Bar{\ts} \mapsto \mathcal{B}(\Bar{\ts} | \ssd_{\ts}, \val^{\ts}_{\df})$ and $\Bar{\ts} \mapsto \mathcal{B}(\ts | \ssd_{\Bar{\ts}}, \val_{\df}^{\ts})$ are continuously differentiable on an open set containing $\ts$. 
\end{assumption}    

\begin{assumption}[Closure under approximate policy improvement] \label{asm:closure}
    For each $\ts \in \Theta$, there exists $\epsilon_{b} \geq 0$ such that 
    \begin{equation*}
        \min_{\ts^+ \in \Theta} \mathcal{B}(\ts^+ | \ssd_{\ts}, \val^{\ts}_{\df}) \leq \min\limits_{\pol \in \Pi} \mathcal{B}(\pol | \ssd_{\ts}, \val^{\ts}_{\df}) + \epsilon_b
    \end{equation*}
     where $\Pi$ is the set of all policies and $\epsilon_b$ is referred to as the inherent Bellman error of the policy class. 
\end{assumption}

\begin{assumption}[Stationary points of the weighted PI objective] \label{asm:stPoint}
    For each $\Bar{\ts} \in \Theta$, the function $\Bar{\ts} \mapsto \mathcal{B}(\Bar{\ts} | \ssd_{\ts}, \val^{\ts}_{\df})$ has no sub-optimal stationary points. 
\end{assumption}

\paragraph{Assumption Verification.} It is easy to see that Assumption~\ref{asm:diff} holds for our policy parameterization in \Cref{eqn:softThresholdPolicy}. Next, we prove that Assumption~\ref{asm:closure} holds for our model in Appendix A using monotonicity arguments. Further, we note that $\epsilon_b \rightarrow 0$ as the sharpness hyperparameter $\g \rightarrow \infty$. This agrees with Lemma 4 in \cite{LinKumar} which shows that the (hard) threshold policy is closed under policy improvement when there are only two servers. Finally, we verify Assumption~\ref{asm:stPoint} by showing that there are indeed no sub-optimal stationary points in Appendix B where we establish convexity of the weighted PI objective.

\begin{figure*}
\centering
\begin{subfigure}{.24\textwidth}
  \centering
  \includegraphics[width=\linewidth]{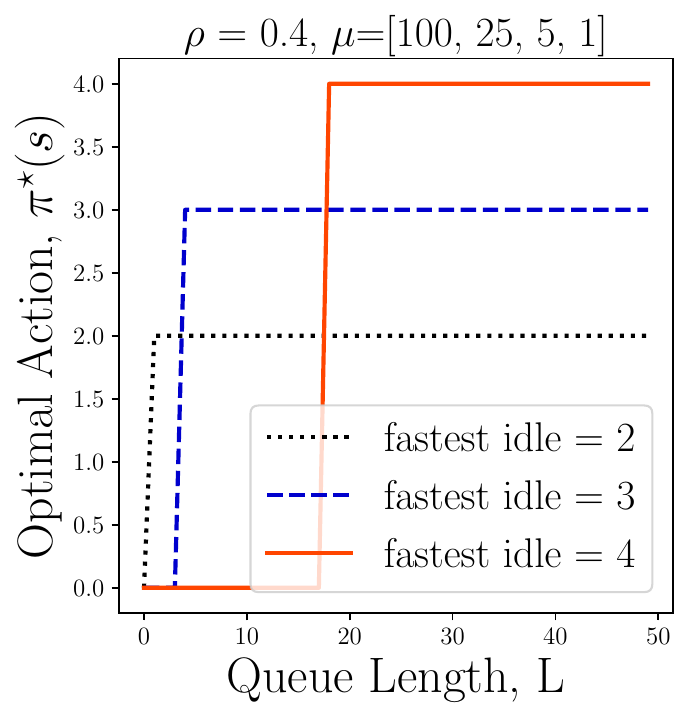}
  \vspace{-10pt}
  \caption{$\srv = [100, 25, 5, 1]$ \\ $\rho = 0.4$}
  \label{fig:E1}
\end{subfigure}%
\begin{subfigure}{.24\textwidth}
  \centering
  \includegraphics[width=\linewidth]{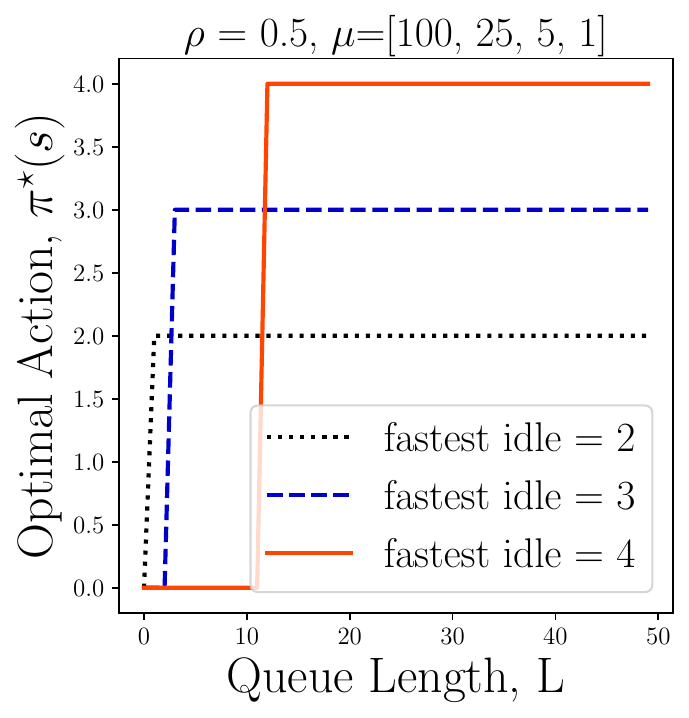}
  \vspace{-10pt}
  \caption{$\srv = [100, 25, 5, 1]$ \\ $\rho = 0.5$}
  \label{fig:E2}
\end{subfigure}%
\begin{subfigure}{.24\textwidth}
  \centering
  \includegraphics[width=\linewidth]{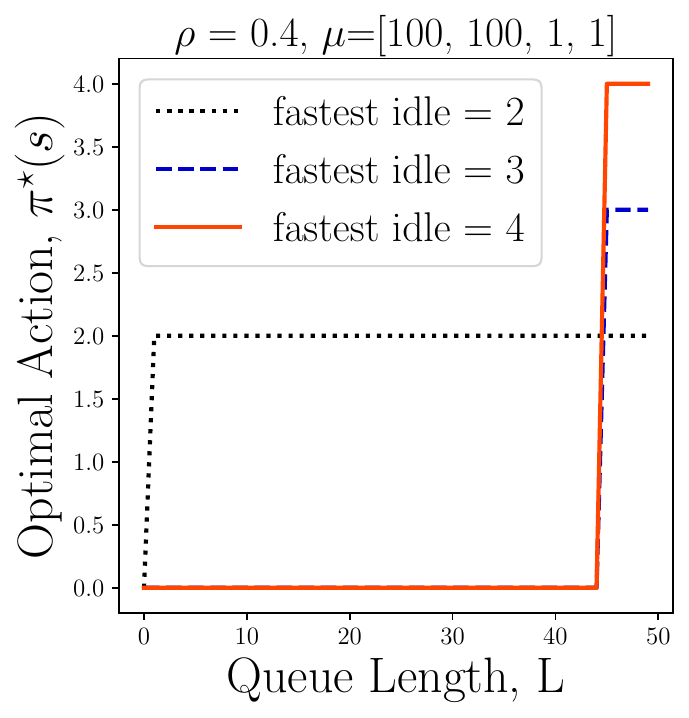}
  \vspace{-10pt}
  \caption{$\srv = [100, 100, 1, 1]$ \\ $\rho = 0.4$}
  \label{fig:E3}
\end{subfigure}%
\begin{subfigure}{.24\textwidth}
  \centering
  \includegraphics[width=\linewidth]{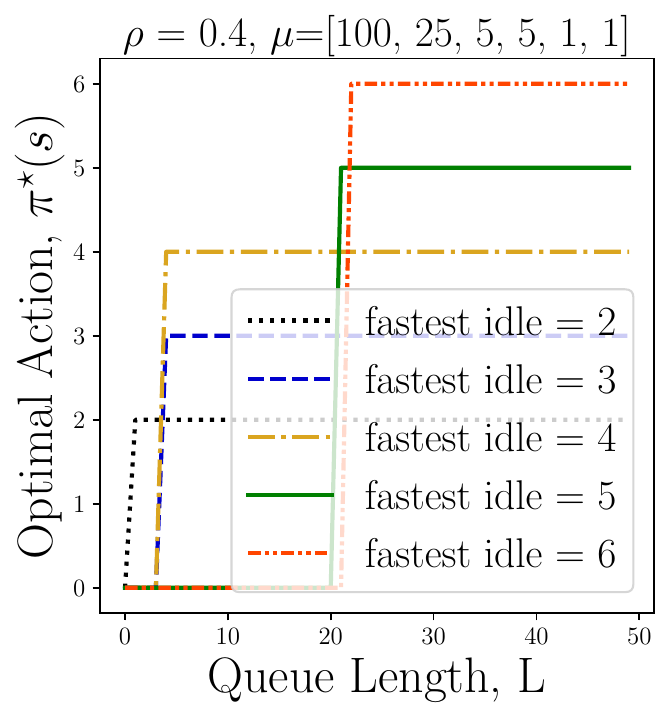}
  \vspace{-10pt}
  \caption{$\srv = [100, 25, 5, 5, 1, 1]$ \\ \centering $\rho = 0.4$}
  \label{fig:E4}
\end{subfigure}
\caption{For multi-server heterogeneous systems, optimal policy is observed to be of threshold type where jobs are routed to the fastest available server only when the queue length exceeds the threshold}
\label{fig:optPol}
\end{figure*}

We now state the formal result of approximate optimality of the stationary point for the two server system in the discounted cost setting.
\begin{theorem} [\cite{BhandariRussoGlobal}, Theorem~5]
    If assumptions 5.6-5.8 hold, then $J_\df$ is continuously differentiable and any stationary point $\ts$ of $J_\df(\cdot)$ satisfies,
    \begin{equation}
        J_{\df}(\pol_{\ts}) - J_{\df}(\pol^\star) \leq \frac{\kappa_{\xi}}{(1-\df)} \cdot \epsilon_b
    \end{equation}
    where $\kappa_{\xi}$ is the effective concentrability coefficient and $\epsilon_b$ is the inherent Bellman error arising due to closure under approximate policy improvement.
\end{theorem}

%%%%%%%%%%%%%%%%%%%%%%

\section{EXPERIMENTS} \label{sec:experiments}

In this section we show empirically that the optimal policy obtained via RVI is of threshold type even for the multi-server case and compare the performance of RVI and \pgName~against two baselines.

\paragraph{Simulation Setup.} We simulate the queueing system as a discrete time system by sampling the continuous time model. If an idle server is imagined to be serving a fictitious job, sampling the continuous time system at instants of arrivals and departures (both true and fictitious) gives us an equivalent discrete time representation \citep{LippmanUniformization}. Note that the standard technique of including fictitious departures ensures that the sample instants are equally exponentially spaced out. We, hence, now have a discrete time system where in at each timestep, whose beginning is marked by an arrival or a departure, the policy decides which server to send a job waiting in the queue to or chooses to keep the job in the queue and wait for the next timestep. The job is then routed accordingly and a transition to the next state occurs as a result of an arrival whose probability is $\ar / (\ar + \sum_i \sri)$ or a departure whose probability is $\sri / (\ar + \sum_i \sri)$.

\paragraph{Baselines.} We compare against two baseline policies --- fastest-available-server (FAS) and ratio-of-service-rate-thresholds (RSRT). FAS routes a job waiting in the queue to the fastest among the available servers at each timestep. RSRT is a threshold policy where a job is routed to the fastest among the available servers (say server $f$) only if the number of jobs waiting in the queue exceeds the threshold $\ts_f = \left(\sum_{i=1}^{f-1} \sri \right) / \sr_f$. The RSRT threshold values, proposed in \cite{Larsen}, represent the maximum number of jobs in the queue beyond which waiting for a faster server is detrimental in a very lightly loaded system ($\ar \rightarrow 0$) where all servers $i<f$ are used. While FAS can be viewed as a pessimistic policy that always routes at the first available opportunity, RSRT is an optimistic policy that believes in the benefit of waiting for a faster server. The goal of our algorithm, thus, is to find an optimal balance between optimism and pessimism.

\begin{figure*}
\centering
\begin{subfigure}{.24\linewidth}
  \centering
  \includegraphics[width=\linewidth]{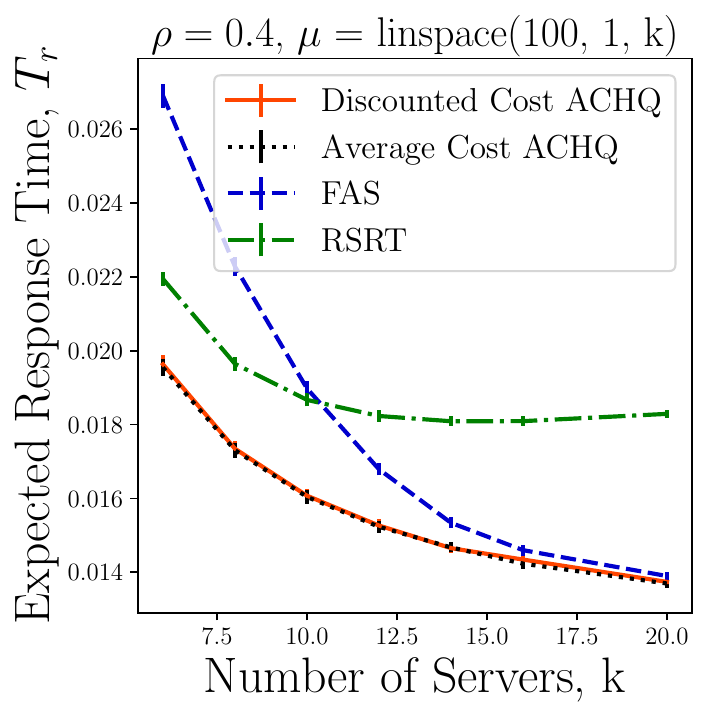}
  % \vspace{-10pt}
  \caption{Number of Servers}
  \label{fig:PGnums}
\end{subfigure}
\begin{subfigure}{.24\linewidth}
  \centering
  \includegraphics[width=\linewidth]{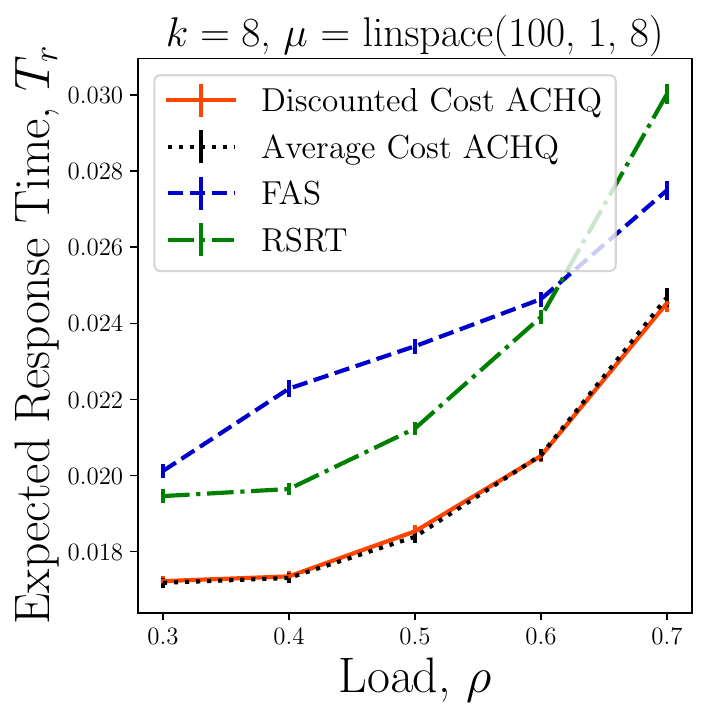}
  % \vspace{-10pt}
  \caption{Load}
  \label{fig:PGload}
\end{subfigure}
\begin{subfigure}{.24\linewidth}
  \centering
  \includegraphics[width=\linewidth]{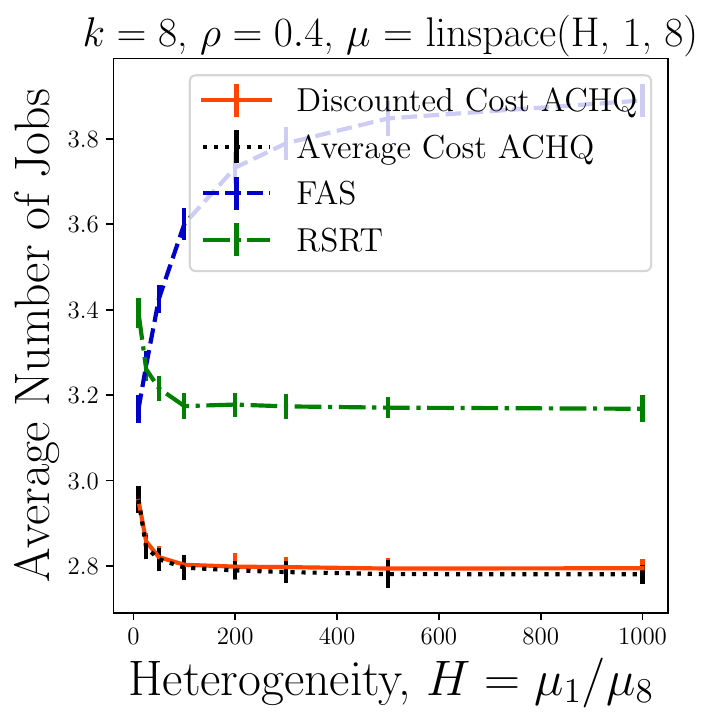}
  % \vspace{-10pt}
  \caption{Heterogeneity}
  \label{fig:PGhet}
\end{subfigure}
\begin{subfigure}{.24\linewidth}
  \centering
  \includegraphics[width=\linewidth]{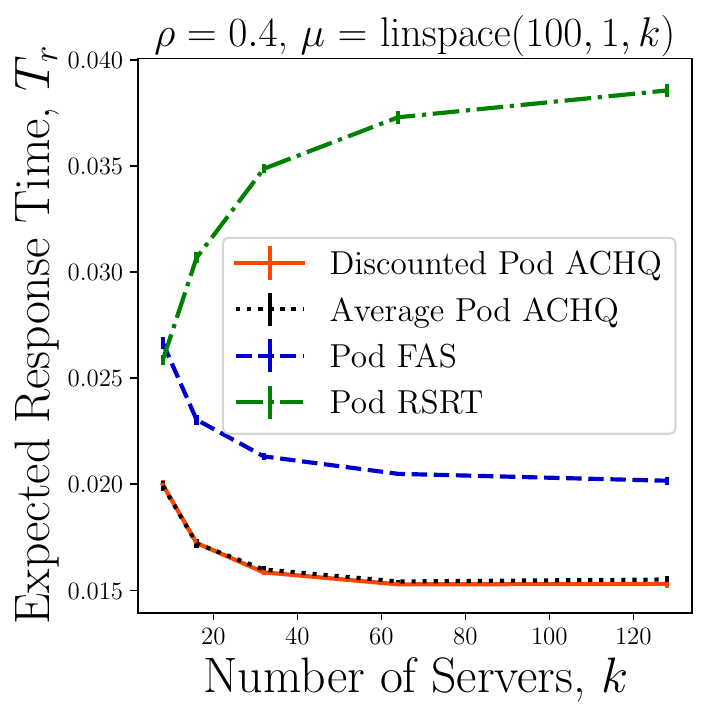}
  % \vspace{-10pt}
  \caption{Pod}
  \label{fig:PoDPG}
\end{subfigure}
\vspace{-5pt}
\caption{\pgName~shows up to $\sim 30\%$ improvement over the FAS and RSRT baselines}
\end{figure*}

\subsection{Relative Value Iteration} \label{sec:expRVI}

In this subsection, we consider four configurations (a)-(d) of the heterogeneous queueing system with a buffer capacity $\buff = 100$ as examples. (a) is the base case where $\srv = [100, 25, 5, 1]$, $\rho = 0.4$. We choose a higher load in (b) where $\srv = [100, 25, 5, 1]$, $\rho = 0.5$. We increase the degree of heterogeneity in (c) and pick $\srv = [100, 100, 1, 1]$, $\rho = 0.4$. We vary the number of servers in (d) with $\srv = [100, 25, 5, 5, 1, 1]$, $\rho = 0.4$. Note that we define load of the system as $\rho = \ar / (\sum_i \sri)$.

\paragraph{Threshold Policy.} Recall that a threshold policy is a rule where a job is sent to the fastest among the available servers (say server $f$) when the number of jobs waiting in the queue is above a threshold $\ts_f$ and otherwise waits for the next timestep. In \Cref{fig:optPol}, we plot the optimal action against the number of jobs waiting in the queue for different states of server occupation. Note that $a=f$ represents the job being routed to the fastest available server $f$ and $a=0$ denotes waiting for the next timestep with the job remaining in the queue. We observe that the optimal policy is indeed of threshold type. Further, an important observation we make is that the faster the service rate of a server, the smaller is the threshold corresponding to it. We notice from (a) and (c) that thresholds of a server are affected by servers faster than it.  Moreover, we see from (a) and (d) that the servers with the same service rate and same set of faster servers have the same threshold despite the number and rates of the slower servers. We also remark that the thresholds depend on load of the system from (a) and (b).

\begin{table}[h]
     \centering
     \begin{tabular}{cccc}
          &  RVI&FAS&RSRT\\ \hline 
          (a)&  5.48 $\pm$ 0.03&  7.72 $\pm$ 0.04&  10.04 $\pm$ 0.05\\ 
          (b)&  8.11 $\pm$ 0.04&  9.72 $\pm$ 0.05&  17.15 $\pm$ 0.08\\ 
          (c)&  2.36 $\pm$ 0.01&  4.64 $\pm$ 0.03&  2.37 $\pm$ 0.01\\ 
          (d)&  5.46 $\pm$ 0.03&  9.56 $\pm$ 0.04&  10.45 $\pm$ 0.06\\ 
     \end{tabular}
     \caption{Expected Response Time, $\ert$ ($\times 10^{-2}$)}
     \vspace{-5pt}
     \label{tab:RVIPerf}
 \end{table}

\paragraph{Performance Improvement.} We observe in \Cref{tab:RVIPerf} that the optimal policy found by Relative Value Iteration (RVI) improves the expected response time by up to $\sim 50\%$ over the FAS baseline. On the other hand, while RVI consistently outperforms RSRT, the amount of gain is highly dependent on the instance. 

\paragraph{Value Function Approximation.} On approximating the value function $V(\stv)$ obtained for the configurations (a) - (d) with a linear function, we obtain $R^2$ values of $0.941, 0.943, 0.942, 0.942$ respectively. This indicates a very high degree of correlation and hence shows that a linear function is a good approximation.
% \vspace{10pt}

\subsection{\pgName} \label{sec:expPG}

\begin{figure}[h]
    \centering
    \includegraphics[width = 0.49\linewidth]{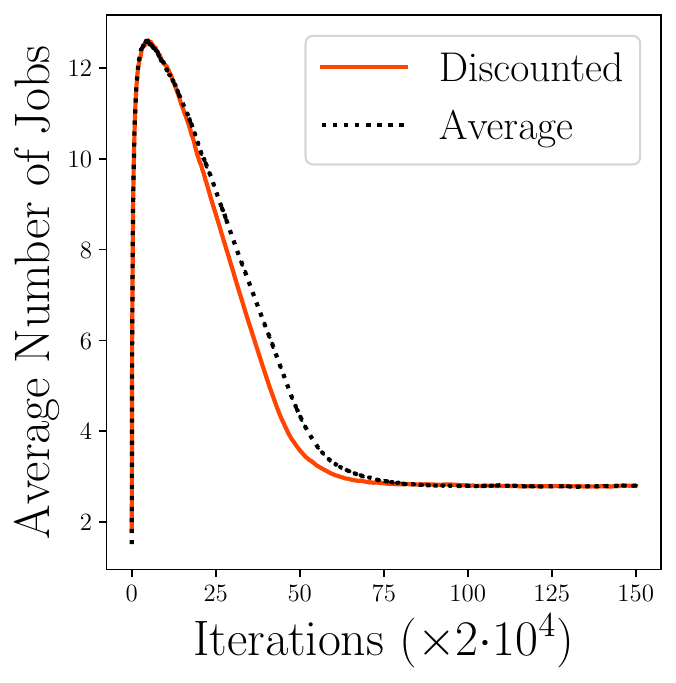}
    \includegraphics[width=0.49\linewidth]{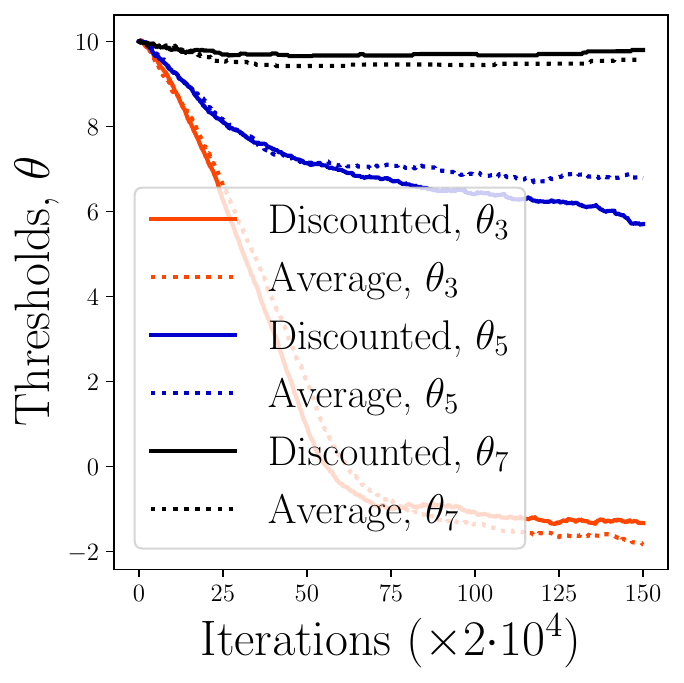}
    \vspace{-15pt}
    \caption{Convergence of \pgName:~Instance of $8$ servers with $\srv = \hbox{linspace}(100, 1)$ and load $\rho = 0.4$}
    \label{fig:PGconverge}
\end{figure}

Here, we consider a representative example of $8$ servers whose service rates are linearly spaced between $100$ and $1$ with a load $\rho = \ar / (\sum_i \sri) = 0.4$ and compare the performance with varying number of server, heterogeneity of service rates and load. We set the sharpness hyperparameter $\g = 1$ and the learning rates for actor $\alpha = 10^{-3}$, critic $\beta = 10^{-3}$ and average cost estimator $\zeta = 10^{-2}$. We observe in \Cref{fig:PGconverge}, that the average number of jobs in the system and the threshold values converge during learning. We see that both the average cost and discount cost versions of the algorithm perform similarly. Note that by Little's Law, the average number of jobs in the system, $n$, and expected response time, $T_r$, are related as $n = \ar T_r$.

With an increase in the \textit{number of servers}, we observe in \Cref{fig:PGnums} that while \pgName~outperforms both the baselines FAS and RSRT, the gap to FAS reduces and gap to RSRT increases. We attribute this to the fact that a moderately fast server is often available as the system scales up. As the \textit{load} increases in \Cref{fig:PGload}, we notice that the gain of \pgName~over FAS increases initially and then decreases. This can be explained by the fact that at high loads even the slower servers are required and at low loads the slower servers are used sparingly. Only at medium loads do we observe non-trivial thresholds and usage of slower servers. With an increase in \textit{heterogeneity} in \Cref{fig:PGhet}, we observe an increased performance gain of \pgName~over FAS. This is because FAS ignores the option of waiting for a increasingly faster server as the heterogeneity increases.

Checking the availability of each server at every timestep becomes expensive as the number of servers in the system scale up. In real world large scale systems, a router often samples a subset of $d$ servers and makes a routing decision based on these. In homogeneous systems, these Power-of-d-Choice (Pod) policies have been shown to be asymptotically optimal \citep{mukherjee2020asymptotic}. We implement a power of $4$ choices version of \pgName~where at each timestep, $4$ servers are sampled without replacement and a job is sent to the fastest among these $4$ servers if the number of jobs in the queue exceeds its threshold. As shown in \Cref{fig:PoDPG}, our algorithm continues to outperform FAS and RSRT. Note that unlike in the case where we are allowed to observe the state of all servers, the gap between FAS and \pgName~does not decrease.

\textit{Remark.} While we consider that the rates are known in this work, \pgName~can easily be extended to the case of unknown arrival and/or service rates by incorporating a simple upper confidence bound based estimation step. 

%%%%%%%%%%%%%%%%%%%%%%

\section{CONCLUSION} \label{sec:conclusion}
We consider the open problem of routing in heterogeneous multi-server queueing systems. We propose \pgName, an efficient policy gradient based algorithm where we leverage the underlying queueing structure by designing a low dimensional soft threshold parameterized policy. We provide stationary-point convergence guarantees for the general case and convergence to an approximate global optimum for the special case of two servers. We also demonstrate an improvement in the expected response time of up to $\sim 30\%$ over the greedy policy of routing to the fastest available server. Directions of future work include proving that the optimal policy is indeed of threshold type even in the multi-server case and expanding to other distributions of arrival and service rates. 

%%%%%%%%%%%%%%%%%%%%%%

\subsubsection*{Acknowledgements}
This work was supported in part by the CMU Dean's Fellowship, Navneet Foundation Endowed Fellowship, CMU CyLab Seed Fund, C3 AI Institute Fund, and NSF grants 2154171, 2339112, ECCS-2145713, CCF-2045694, CNS-2112471, CPS-2111751, ONR-N00014-23-1-2149.

\pagebreak
% \subsubsection*{References}
\bibliography{references}

%%%%%%%%%%%%%%%%%%%%%%%%%%%%%%%%%%%%%%%%%%%%%%%%%%%%%%%%%%%%
\section*{Checklist}

% %%% BEGIN INSTRUCTIONS %%%
% The checklist follows the references. For each question, choose your answer from the three possible options: Yes, No, Not Applicable.  You are encouraged to include a justification to your answer, either by referencing the appropriate section of your paper or providing a brief inline description (1-2 sentences). 
% Please do not modify the questions.  Note that the Checklist section does not count towards the page limit. Not including the checklist in the first submission won't result in desk rejection, although in such case we will ask you to upload it during the author response period and include it in camera ready (if accepted).

% \textbf{In your paper, please delete this instructions block and only keep the Checklist section heading above along with the questions/answers below.}
% %%% END INSTRUCTIONS %%%

 \begin{enumerate}

 \item For all models and algorithms presented, check if you include:
 \begin{enumerate}
   \item A clear description of the mathematical setting, assumptions, algorithm, and/or model. [Yes]
   \item An analysis of the properties and complexity (time, space, sample size) of any algorithm. [Yes]
   \item (Optional) Anonymized source code, with specification of all dependencies, including external libraries. [Not Applicable]
 \end{enumerate}

 \item For any theoretical claim, check if you include:
 \begin{enumerate}
   \item Statements of the full set of assumptions of all theoretical results. [Yes]
   \item Complete proofs of all theoretical results. [Yes]
   \item Clear explanations of any assumptions. [Yes]     
 \end{enumerate}

 \item For all figures and tables that present empirical results, check if you include:
 \begin{enumerate}
   \item The code, data, and instructions needed to reproduce the main experimental results (either in the supplemental material or as a URL). [Yes]
   \item All the training details (e.g., data splits, hyperparameters, how they were chosen). [Yes]
    \item A clear definition of the specific measure or statistics and error bars (e.g., with respect to the random seed after running experiments multiple times). [Yes]
    \item A description of the computing infrastructure used. (e.g., type of GPUs, internal cluster, or cloud provider). [Not Applicable]
 \end{enumerate}

 \item If you are using existing assets (e.g., code, data, models) or curating/releasing new assets, check if you include:
 \begin{enumerate}
   \item Citations of the creator If your work uses existing assets. [Not Applicable]
   \item The license information of the assets, if applicable. [Not Applicable]
   \item New assets either in the supplemental material or as a URL, if applicable. [Not Applicable]
   \item Information about consent from data providers/curators. [Not Applicable]
   \item Discussion of sensible content if applicable, e.g., personally identifiable information or offensive content. [Not Applicable]
 \end{enumerate}

 \item If you used crowdsourcing or conducted research with human subjects, check if you include:
 \begin{enumerate}
   \item The full text of instructions given to participants and screenshots. [Not Applicable]
   \item Descriptions of potential participant risks, with links to Institutional Review Board (IRB) approvals if applicable. [Not Applicable]
   \item The estimated hourly wage paid to participants and the total amount spent on participant compensation. [Not Applicable]
 \end{enumerate}

 \end{enumerate}

%%%%%%%%%%%%%%%%%%%%%%%%%%%
\newpage
\onecolumn
\appendix

\section{CLOSURE UNDER APPROXIMATE POLICY IMPROVEMENT}

In this section, we verify that Assumption~5.7 holds for our problem by presenting a proof of closure of the soft threshold policy under approximate policy improvement. This follows the arguments made in Lemma~4 of \cite{LinKumar} which shows that (hard) threshold policies are closed under policy improvement for the two server case.

Recall that the queue is represented as a discrete time system where at each timestep, whose beginning is marked by an arrival or a departure, the router decides if and which server to send a job waiting in the queue to. We consider without loss of generality that the arrival and service rates are normalized as $\ar + \sum_r \sri = 1$. 
% Also, recollect that the Bellman equation with discount factor $\df$ can be written as $\val^{\pol}_{\df}(\stv) = c(\stv) + \df \sum\limits_{\stv' \in \mathcal{S}} \prob(\stv' | \stv, a)\val^{\pol}_{\df}(\stv)$ and the weighted Policy Iteration (PI) objective as $\mathcal{B}(\Bar{\pol} | \ssd, \val^{\pol}_\df) = \mathbb{E}_{\stv \sim \ssd}[(T_{\Bar{\pol}}\val^\pol_\df)(\stv)]$. 
The soft threshold policy parameterized by $\ts \in \Theta$, can be written as 
% \neharika{Change this to continuous time representation where even two jobs are allowed to be scheduled at the same time}
\begin{align}
    \pol_{\ts} (a = 0 | \stv=(\q, 1, 1)) & = 1 \nonumber \\
    \nonumber\\
    \pol_{\ts} (a = 1 | \stv=(\q, 0, 1)) & = 1 \hspace{10pt} \mbox{if } \q > 0 \nonumber \\
    \pol_{\ts} (a = 0 | \stv=(\q, 0, 1)) & = 1 \hspace{10pt} \mbox{if } \q = 0 \nonumber \\
    \nonumber\\
    \pol_{\ts}(a = 2 | \stv = (\q, 1, 0)) & = \frac{e^{\g (\q - \ts)}}{1 + e^{\g (\q - \ts)}} \nonumber \\
    \pol_{\ts}(a = 0 | \stv = (\q, 1, 0)) & = \frac{1}{1 + e^{\g (\q - \ts)}} \nonumber \\
    \nonumber\\
    \pol_{\ts}(a = 1 | \stv = (\q, 0, 0)) & = 1 \hspace{10pt} \mbox{if } \q > 0 \nonumber \\
    \pol_{\ts}(a = 0 | \stv = (\q, 0, 0)) & = 1 \hspace{10pt} \mbox{if } \q = 0
\end{align}
where $\pol_{\ts}(a | \stv)$ represents the probability that a router following policy $\pol_{\ts}$ takes action $a$.

For a given soft threshold policy $\pol_{\ts}$, represent the policy obtained by policy improvement over the set of all policies as $\pol' = \argmin\limits_{\pol \in \Pi} \mathcal{B}(\pol | \ssd_{\ts}, \val^{\ts}_{\df})$. Similarly, denote the policy obtained by policy improvement over the set of soft threshold policies as $\ts' = \argmin\limits_{\ts^+ \in \Theta} \mathcal{B}(\ts^+ | \ssd_{\ts}, \val^{\ts}_{\df})$. To verify Assumption~5.7 for our problem, we want to show that the weighted PI objective of $\ts'$ is approximately equal to that of $\pol'$. Formally, we want to show that there exists a $\ts' \in \Theta$ for every soft threshold policy $\ts \in \Theta$ such that $$\mathcal{B}(\ts' | \ssd_{\ts}, \val^{\ts}_{\df}) \leq \mathcal{B}(\pol' | \ssd_{\ts}, \val^{\ts}_{\df}) + \epsilon_b$$ where $\epsilon_b$ is referred to as the inherent Bellman error. 

We first consider the improvement over the set of all policies $\pol'$. Define
\begin{align}
    h_0 & = \val^{\ts}_{\df}(0, 1, 0) - \val^{\ts}_{\df}(0, 0, 1) \nonumber \\
    h_{\q} & = \val^{\ts}_{\df}(\q, 1, 0) - \val^{\ts}_{\df}(\q-1, 1, 1) \hspace{10pt} \mbox{for } \q \geq 1.
\end{align}
For state $\stv=(\q, 1, 0)$, notice that $h_\q < 0$ implies that the optimal action under $\pol'$ is to wait until the next timestep and $h_\q \geq 0$ implies that the optimal action is to route a job to the slow server. If $h_\q < 0$ for $\q \leq \q^\star$ and $h_\q \geq 0$ for $\q > \q^\star$, then the improved policy $\pol'$ is a threshold policy where a job is routed to the slow server only when the number of jobs in the queue exceeds the threshold $\q^\star$ (see Lemma~4, \cite{LinKumar}). 

Now, if we pick $\ts' = \q^\star$, the weighted PI objective of improved soft threshold policy $\ts'$ will approximately be equal to that of $\pol'$ with $\epsilon_b$ characterizing the error. The Bellman approximation error $\epsilon_b$ arises due to the difference between $\pol'$ being a (hard) threshold policy and $\ts'$ being a soft threshold policy. Note that $\epsilon_b \rightarrow 0$ as the decision sharpness hyperparameter $\g \rightarrow \infty$ and the soft threshold tends towards a hard threshold. 

We dedicate the rest of the section to proving that the structure required over $h_\q$ where $h_\q < 0$ for $\q \leq \q^\star$ and $h_\q \geq 0$ for $\q > \q^\star$ is indeed true using monotonicity arguments.

Applying Bellman equation to $\val^{\ts}_{\df}(\q, 1, 0)$ and $\val^{\ts}_{\df}(\q-1, 1, 1)$, we have
\begin{align}
    \val^{\ts}_{\df}(\q, 1, 0) = (\q + 1) & + \df\ar\left[\frac{e^{\g(\q+1-\ts)}}{1 + e^{\g(\q+1-\ts)}}\val^{\ts}_{\df}(\q, 1, 1) + \frac{1}{1+e^{\g(\q+1-\ts)}}\val^{\ts}_{\df}(\q+1, 1, 0) \right] + \df\sr_1\val_\df^\ts(\q-1, 1, 0) \nonumber\\
    % & + \df\sr_1\left[\frac{e^{\g(\q-1-\ts)}}{1+e^{\g(\q-1-\ts)}}\val^{\pol}_{\df}(\q-2, 1, 1) + \frac{1}{1+e^{\g(\q-1-\ts)}}\val^{\pol}_{\df}(\q-1, 1, 0)\right] \nonumber \\
    & + \df\sr_2\left[\frac{e^{\g(\q-\ts)}}{1+e^{\g(\q-\ts)}}\val^{\ts}_{\df}(\q-1, 1, 1) + \frac{1}{1+e^{\g(\q-\ts)}}\val^{\ts}_{\df}(\q, 1, 0)\right], \\
    \val^{\ts}_{\df}(\q-1, 1, 1) = (l+1) & + \df\ar\val^{\ts}_{\df}(\q, 1, 1) + \df\sr_1\val^{\ts}_{\df}(\q-2, 1, 1) \nonumber \\
    & + \df\sr_2\left[\frac{e^{\g(\q-1-\ts)}}{1+e^{\g(\q-1-\ts)}}\val^{\ts}_{\df}(\q-2, 1, 1) + \frac{1}{1+e^{\g(\q-1-\ts)}}\val^{\ts}_{\df}(\q-1, 1, 0) \right].
\end{align}
Subtracting the two, we get, 
\begin{align} \label{eqn:hSub}
    h_\q & = \val^{\ts}_{\df}(\q, 1, 0) - \val^{\ts}_{\df}(\q-1, 1, 1) \nonumber \\
    & = \df\ar\left[\frac{\val^{\ts}_{\df}(\q+1, 1, 0) - \val^{\ts}_{\df}(\q, 1, 1)}{1+e^{\g(\q+1-\ts)}} \right] + \df\sr_1 \left[\val^{\ts}_{\df}(\q-1, 1, 0) - \val^{\ts}_{\df}(\q-2, 1, 1) \right] \nonumber \\
    % & = \df\ar\left[\frac{\val^{\pol}_{\df}(\q+1, 1, 0) - \val^{\pol}_{\df}(\q, 1, 1)}{1+e^{\g(\q+1-\ts)}} \right] + \df\sr_1 \left[\frac{\val^{\pol}_{\df}(\q-1, 1, 0) - \val^{\pol}_{\df}(\q-2, 1, 1)}{1+e^{\g(\q-1-\ts)}} \right] \nonumber \\
    & + \df\sr_2 \bigg[\frac{e^{\g(\q-\ts)}}{1+e^{\g(\q-\ts)}}\val^{\ts}_{\df}(\q-1, 1, 1) + \frac{1}{1+e^{\g(\q-\ts)}}\val^{\ts}_{\df}(\q, 1, 0) \nonumber\\
    & - \frac{e^{\g(\q-1-\ts)}}{1+e^{\g(\q-1-\ts)}}\val^{\ts}_{\df}(\q-2, 1, 1) - \frac{1}{1+e^{\g(\q-1-\ts)}}\val^{\ts}_{\df}(\q-1, 1, 0)\bigg].
\end{align}

Now, we argue that for a large enough sharpness hyperparameter $\g$, there exists an $\q_1 > 0$ such that $h_\q \geq 0$ for all $\q > \q_1$ and denote the smallest such $\q_1$ as $\q^\star$. This can be seen by the fact that terms 1 and 3 corresponding to a departure from the second server dominate the equation at large queue lengths and we know $$\frac{e^{\g(\q-\ts)}}{1+e^{\g(\q-\ts)}}\val^{\ts}_{\df}(\q-1, 1, 1) \geq \frac{e^{\g(\q-1-\ts)}}{1+e^{\g(\q-1-\ts)}}\val^{\ts}_{\df}(\q-2, 1, 1).$$ This is because $\val^{\ts}_{\df}(\q-1, 1, 1) \geq \val^{\ts}_{\df}(\q-2, 1, 1)$ following arguments of monotonicity of Bellman operator as in Lemma 1 of \cite{LinKumar}.

With some algebraic manipulation, we obtain,
\begin{align}
    (1-b\df)h_\q & \geq b_1\df\ar(h_{\q+1} - h_{\q}) + b_2\df\sr_1(h_{\q-1}-h_\q) \nonumber \\
    & + \frac{\df\sr_2}{1+e^{\g(\q-1-\ts)}}\left[\frac{1+e^{\g(\q-1-\ts)}}{1+e^{\g(\q-\ts)}}\val^{\ts}_{\df}(\q-1, 1, 1) - \val^{\ts}_{\df}(\q-1, 1, 0) \right]
\end{align}
where $b_1 = 1/(1+e^{\g(\q+1-\ts)})$, $b_2 = 1$,
% $ b_2 = 1/(1+e^{\g(\q-1-\ts)})$, 
$ b_3 = 1/(1+e^{\g(\q-\ts)})$ and $b = b_1\ar + b_2\sr_1 + b_3\sr_2$.
For a large enough sharpness constant $\g$, the third term on the right hand side in the expression above is always positive. We hence have 
\begin{equation}\label{eqn:hIneq}
    -(1-a\df)h_\q + \df\ar(h_{\q+1}-h_\q) \leq \df\sr_1(h_\q - h_{\q-1}).
\end{equation}
We know $h_{\q^\star+1} \geq 0$ and $h_{\q^\star} < 0$. Using $\q = \q^\star, \q^\star-1, \dots$ in \Cref{eqn:hIneq}, it follows that 
\begin{equation} \label{eqn:hNegOrder}
    h_0 < h_1 < \dots h_{\q^\star} < 0.
\end{equation}

Putting \Cref{eqn:hSub} and \Cref{eqn:hNegOrder} together, we have $h_0 < h_1 < \dots h_{\q^\star} < 0$ and $h_{\q} \geq 0$ for all $\q > \q^\star$. The improved policy $\pol'$ is thus a threshold policy with threshold $\q^\star$. Choosing $\ts' = \q^\star$ results in approximate equality of the weighted PI objectives as detailed above and concludes the proof.

%%%%%%%%%%%%%%%%%%%%%%%%%%

\newpage
\section{STATIONARY POINTS OF WEIGHTED PI OBJECTIVE}

In this section, we show that the weighted PI objective has no sub-optimal stationary points by arguing that it is convex. Let $l^\star$ be such that $Q^\ts_\df((\q, 1, 0), a=2) \leq Q^\ts_\df((\q, 1, 0), a=0)$ for all $\q > l^\star$ and $Q^\ts_\df((\q, 1, 0), a=2) > Q^\ts_\df((\q, 1, 0), a=0)$ for all $\q \leq \q^\star$. The existence of such an $\q^\star$ is guaranteed by \Cref{eqn:hSub} in Appendix A above. Now, consider the value of the weighted PI objective $\mathcal{B}(\Bar{\ts}|\ssd_\ts, \val^\ts_\df)$ with increasing values $\Bar{\ts}$. As $\Bar{\ts}$ increases towards $\q^\star$, the number of states where the sub-optimal action of routing a job to the slow server reduces and hence the weighted PI objective monotonically reduces. The minimum is obtained at $\Bar{\ts}=\q^\star$. Further, as $\Bar{\ts}$ increases beyond $\q^\star$, the number of states where the sub-optimal action of idling until the next timestep is chosen more often resulting in a monotonic increase in the weighted PI objective. Thus, the weighted PI objective is a convex function in $\Bar{\ts}$ and there are no sub-optimal stationary points.

\section{RELATIVE VALUE ITERATION}
We detail the Relative Value Iteration algorithm below for the sake for completeness. Note that the span semi-norm used in the algorithm is defined as $sp(f(x)) := [\max_x f(x)][\min_x f(x)]$. 
% \neharika{include a note on the correctness/convergence of this algorithm?}

\begin{algorithm}[H]
    \caption{Relative Value Iteration} \label{algo:RVI}
    \begin{algorithmic}[1]
         \STATE \textbf{Initialize} $V^{(0)}(\stv) = 0$ $\forall \stv \in \mathcal{S}$, $\epsilon$
         \WHILE{$sp(\val^{(t)} - \val^{(t-1)}) > \epsilon$}
             \FOR{$\stv \in \mathcal{S}$}
                \STATE $\val^{(t+1)}(\stv) \gets T(\val^{(t)})(\stv) - T(\val^{(t)})(\stv^\star)$ \label{line:Vupdate}
             \ENDFOR
             \STATE $t \gets t+1$
         \ENDWHILE
         \STATE $\pol(\stv) = \argmin\limits_a c(\stv) + \sum_{\stv' \in \mathcal{S}} \prob(\stv' \lvert \stv, a)\val^{(t)}(\stv')$
    \end{algorithmic}
\end{algorithm}

\end{document}